\documentclass[
]{ceurart}

\usepackage{listings}
\begin{document}

\copyrightyear{2025}
\copyrightclause{Copyright for this paper by its authors.
  Use permitted under Creative Commons License Attribution 4.0
  International (CC BY 4.0).}

\conference{Workshop SIG Knowledge Management (FG WM) on 16th September 2025 at KI 2025, Potsdam, Germany}

\title{Tripartite-GraphRAG via Plugin Ontologies}

%\tnotetext[1]{You can use this document as the template for preparing your
%  publication. We recommend using the latest version of the ceurart style.}

\author{Michael Banf}[%
email=michael.banf@perelyn.com,
]
\cormark[1]

\author{Johannes Kuhn}[%
email=johannes.kuhn@perelyn.com,
]

\address{Perelyn GmbH, Reichenbachstraße 31, 80469 München, Germany }

\cortext[1]{Corresponding author.}

\begin{abstract}
Large Language Models (LLMs) have shown remarkable capabilities across various domains, yet they struggle with knowledge-intensive tasks in areas that demand factual accuracy, such as in industrial automation and healthcare. Key limitations include their tendency to hallucinate, lack of source traceability (provenance), and challenges in timely knowledge updates. Retrieval Augmented Generation (RAG) techniques have attempted to address these issues by incorporating external knowledge, but they face their own limitations, chiefly the reliance on embedding-based similarity for textual chunk selection, which can be unreliable or lack coverage due to fluctuating chunk sizes and topic diversity within queries. As a result, these methods often produce incomplete, noisy, or sub-optimally arranged input prompts for the LLM, especially for complex, multi-aspect queries. Combining language models with knowledge graphs (GraphRAG) offers promising avenues for overcoming these deficits. However, a major challenge lies in creating such a knowledge graph in the first place. Here, we report on our ongoing work and first experiences combining LLMs with a novel tripartite knowledge graph representation. The graph is constructed by connecting complex, domain-specific objects via a curated ontology of corresponding, domain-specific concepts to relevant sections within chunks of text through a concept-anchored pre-analysis of source documents starting from an initial lexical graph. Subsequently, we formulate LLM prompt creation as an unsupervised node classification problem allowing for the optimization of information density, coverage, and arrangement of LLM prompts at significantly reduced lengths. An initial experimental evaluation of our approach on a healthcare use case, involving multi-faceted analyses of patient anamneses given a set of medical concepts as well as a series of clinical guideline literature, indicates its potential to optimize information density, coverage, and arrangement of LLM prompts while significantly reducing their lengths, which, in turn, may lead to reduced costs as well as more consistent and reliable LLM outputs.
\end{abstract}


\maketitle

\section{Introduction}

While language models (LLMs) have demonstrated impressive capabilities, they still have their limitations in knowledge-intensive tasks - especially in areas where factually correct information is essential, such as industrial automation or healthcare. These weaknesses can be traced back to fundamental challenges: Firstly, LLMs are prone to hallucination, i.e. generating statements that are incorrect or misleading in terms of content. Secondly, their updatability is limited - it is difficult for them to integrate new knowledge promptly or to adapt to dynamic information environments. Another central problem is the lack of provenance, i.e. the lack of traceability of the information source, which makes it difficult to assess the trustworthiness of its output. 

To this end, Retrieval-Augmented Generation (RAG) has emerged as a powerful paradigm for enhancing large language models with external knowledge~\cite{lewis2020retrieval}. Recent advances have focused on scaling and efficiency~\cite{borgeaud2022improving}, dynamically determining retrieval needs~\cite{jiang2023active}, or tackling long-context challenges through re-ranking mechanisms and token-level retrieval respectively~\cite{xu2023retrieval, ram2023context}.

As a next step, the combination of language models with knowledge graphs represents a significant evolution by incorporating graph-structured knowledge ~\cite{edge2024local} demonstrating how knowledge graphs enable more structured retrieval for complex relationships and multi-hop reasoning. The integration of graph neural networks has shown promise, with joint reasoning over text and knowledge graphs~\cite{yasunaga2022deep}, textual graph understanding~\cite{he2024g}, or hierarchical frameworks for multi-level graph abstraction~\cite{zhang2024graphrag}.

However, a major challenge lies in creating such a knowledge graph in the first place - on the basis of suitable data sources, expandable and in line with the respective application objective. To this end, methodologies range from low granularity, layout preserving parsing and hierarchical representation of document structures, which allow for the traversal of document trees during retrieval beyond relying purely on semantic similarities~\cite{Yu2024Neo4j}, to the fine-grained extraction and clustering of entities and relationships from text chunks, including all associated obstacles including costs, extendability, or entity resolution and deduplication challenges~\cite{edge2025localglobalgraphrag, Neo4j2025}. 

Here, we report on our ongoing work to propose a complementary perspective through the construction of a tripartite knowledge graph by connecting complex, domain-specific objects as a 'plugin ontology' of corresponding, domain-specific concepts to relevant sections within chunks of text through a concept-anchored pre-analysis of source documents starting from an initial lexical graph. As a consequence, our Tripartite-GraphRAG:

\begin{itemize}
\item implements a concept-specific, information-preserving pre-compression of textual chunks
\item allows for the formation of a concept-specific relevance estimation of embedding similarities grounded in statistics
\item avoids common challenges w.r.t. continuous extendability, such as the need for entity resolution and deduplication with the addition of new objects, concepts or documents.
\end{itemize}

Finally, we formulate LLM prompt creation as a node classification problem allowing for optimized information density, coverage, and arrangement of LLM prompts at significantly reduced lengths.

\begin{figure}
\includegraphics[width=\textwidth]{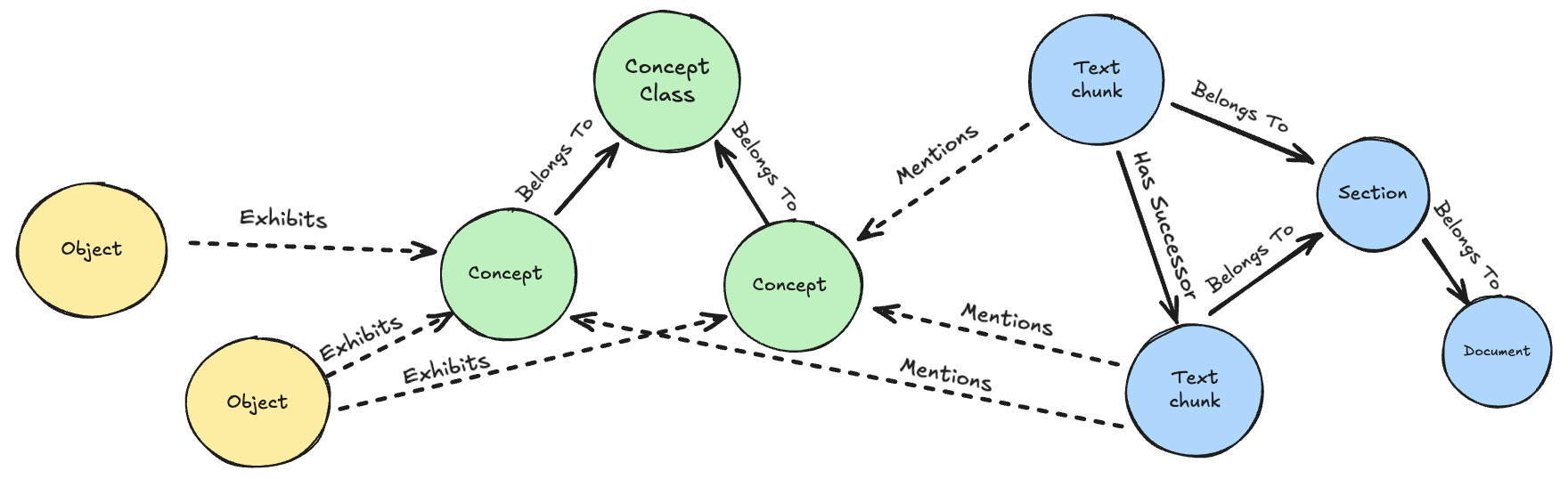}
\caption{A tripartite knowledge graph representation connecting complex, domain-specific objects of investigation (yellow) via a curated ontology of corresponding, domain-specific concepts (green) to relevant sections within chunks of text (blue) through a concept-anchored pre-analysis of source documents.} \label{fig_tripartite_perspective}
\end{figure}

\section{Methodology}

\subsection{Information preserving, concept-specific summarizations via plugin ontologies}

Given a set of documents, we create an initial lexical graph as described in ~\cite{Yu2024Neo4j}. In order to map their adjacent nature we introduce successor-type relationships between text chunks $t$. Next, we curate and plugin an ontology of domain-specific concept class and corresponding concept $c$ nodes, such as clinical terminology (see experiments). With respect to $c$, at the current stage of development, we demonstrate the methodology via adhoc, i.e. manually curated, ontologies. Rationale of the manual ontology creation is that the conceptual design of the tripartite graph as supporting structure is invariant to the way the ontology is created. However, the approach is supposed to plugin established domain-specific ontologies to enhance advanced domain-specific reasoning as well (see discussion).

\begin{figure}[t!]
\includegraphics[width=\textwidth]{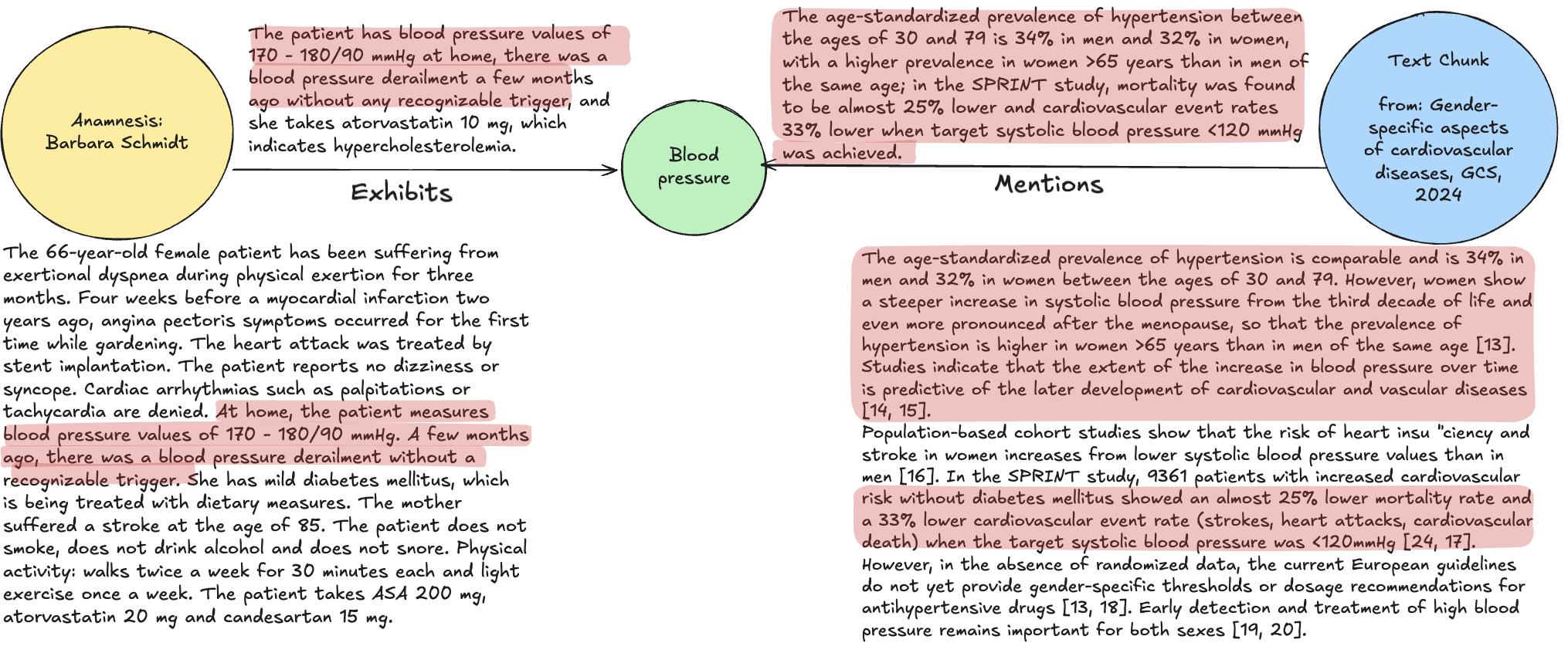}
\caption{Left: object of investigation $o$ (yellow), here fictitious patient anamnesis, and information-preserving pre-compression of $o$ w.r.t concept $c$ (green, $c = \text{blood pressure}$). Right: information-preserving pre-compression of a text chunk $t$ (blue) w.r.t. $c$, i.e. retaining relevant information (red) while discarding most unrelated text.} \label{fig_concept_specific_summaries}
\end{figure}

Further, using very basic prompting, we task an LLM, per concept $c$, to assess each text chunk $t$ regarding any relevant information in connection with $c$, and, in case, create a corresponding relationship $e_{c,t}$, while storing extracted information as edge property in $e_{c,t}$ (see figure~\ref{fig_concept_specific_summaries}). As a consequence, our approach implements a concept-specific, information-preserving pre-compression of textual chunks $t$ w.r.t. $c$, i.e. retaining and storing relevant information while discarding as much textual noise as possible for each unique relationship $e_{c,t}$. This is in contrast to more general, i.e. non-concept guided, pre-summarizations of text chunks at knowledge graph construction time potentially leading to information loss w.r.t a later query. Finally, we introduce a third type of node, which represents complex, domain-specific objects of investigation $o$, such as a patient's medical anamnesis, involving multiple relevant elements for cross-examination (see figure~\ref{fig_concept_specific_summaries}). Again, we task the LLM to scan all objects $o$ regarding any connection to any of the concepts $c$ and, in turn, create relationships $e_{o,c}$. The resulting graph schema is visualized in figure~\ref{fig_tripartite_perspective}. Note that we allow violations of the strict tripartite graph definition w.r.t. allowing for concept and document structure nodes to be adjacent or hierarchically connected. Further note, that our simple two-level concept ontology similarily generalizes to deeper concept hierarchies.

\subsection{Prompt creation as node classification problem}

\begin{figure}[t!]
\centering
\includegraphics[width=0.8\textwidth]{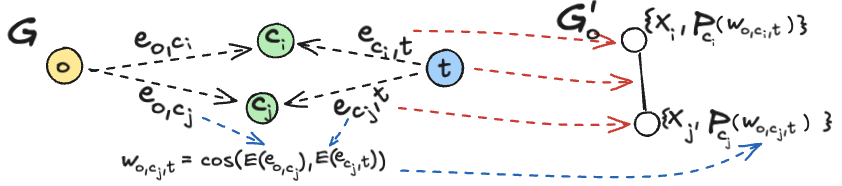}
\caption{Illustration of graph transformation of $G$ related to object $o$ to create classification graph $G'_o$ that represents the information in each $e_{c,t}$ between concept $c$ and text chunk $t$ related to $o$, measured as $w_{o,c,t}$ in relation to $c$'s specific distribution of similarity scores $P_{c}$. Here $E_{e_{o,c}}, E_{e_{c,t}}$ denote the vector embeddings of the information stored in $e_{o,c}$ and $e_{c,t}$ with $cos(.)$ denoting the cosine similarity.} \label{fig_graph_transformation}
\end{figure}

A major challenge of common RAG approaches is the selection of appropriate text chunks w.r.t a given query, chiefly because of the reliance on embedding-based similarity for textual chunk selection, which can be unreliable or lack coverage due to fluctuating chunk sizes and topic diversity within queries. To this end, our approach provides a natural remedy. Applying a transformation to the constructed knowledge graph, we formulate LLM prompt creation as an unsupervised node classification problem (see illustration in figure~\ref{fig_graph_transformation}). Thus, we apply a transformation to parts of our tripartite knowledge graph $G$ related to an object of investigation $o$ to create a novel classification graph $G'_o$ that represents the information in each relationship $e_{c,t}$ between $c$ and $t$ related to $o$ - i.e. given a specific relationship $e_{o,c}$ between $o$ and $c$ - as a node $x \in 0, 1$. 

Here, $x$ in $G'_o$ describes the state of the information stored in $e_{c,t}$ to be or not to be included in the prompt related to a query on $o$, i.e. $x = 1$ or $0$ otherwise. The state of $x$ is governed by the (semantic) relevance of the stored information in $e_{c,t}$ related to $o$'s association with $c$, i.e. $e_{o,c}$ that can be measured as $w_{o,c,t} = cos(E_{e_{o,c}}, E_{e_{c,t}})$. Here $E_{e_{o,c}}, E_{e_{c,t}}$ denote the vector embeddings of the information stored in $e_{o,c}$ and $e_{c,t}$ with $cos(.)$ denoting the cosine similarity. By collecting all $w_{o,c,t}$ for all $o$ and $t$ connected to a specific concept $c$, we derive an empirical probability distribution $P_{c}$ per each $c$ for all observed similarity scores $w_{o,c,t}$ w.r.t. $c$. Thus, $x$ can be assigned a probability $P_c(w_{o,c,t})$ w.r.t. the concept $c$'s specific $P_{c}$. If $P_c(w_{o,c,t}) > \alpha$ then $x = 1$ with $\alpha$ denoting some hyper-parameter on $P_c$, e.g. $\alpha = 0.9$. 

\begin{figure}[b!]
\centering
\includegraphics[width=1\textwidth]{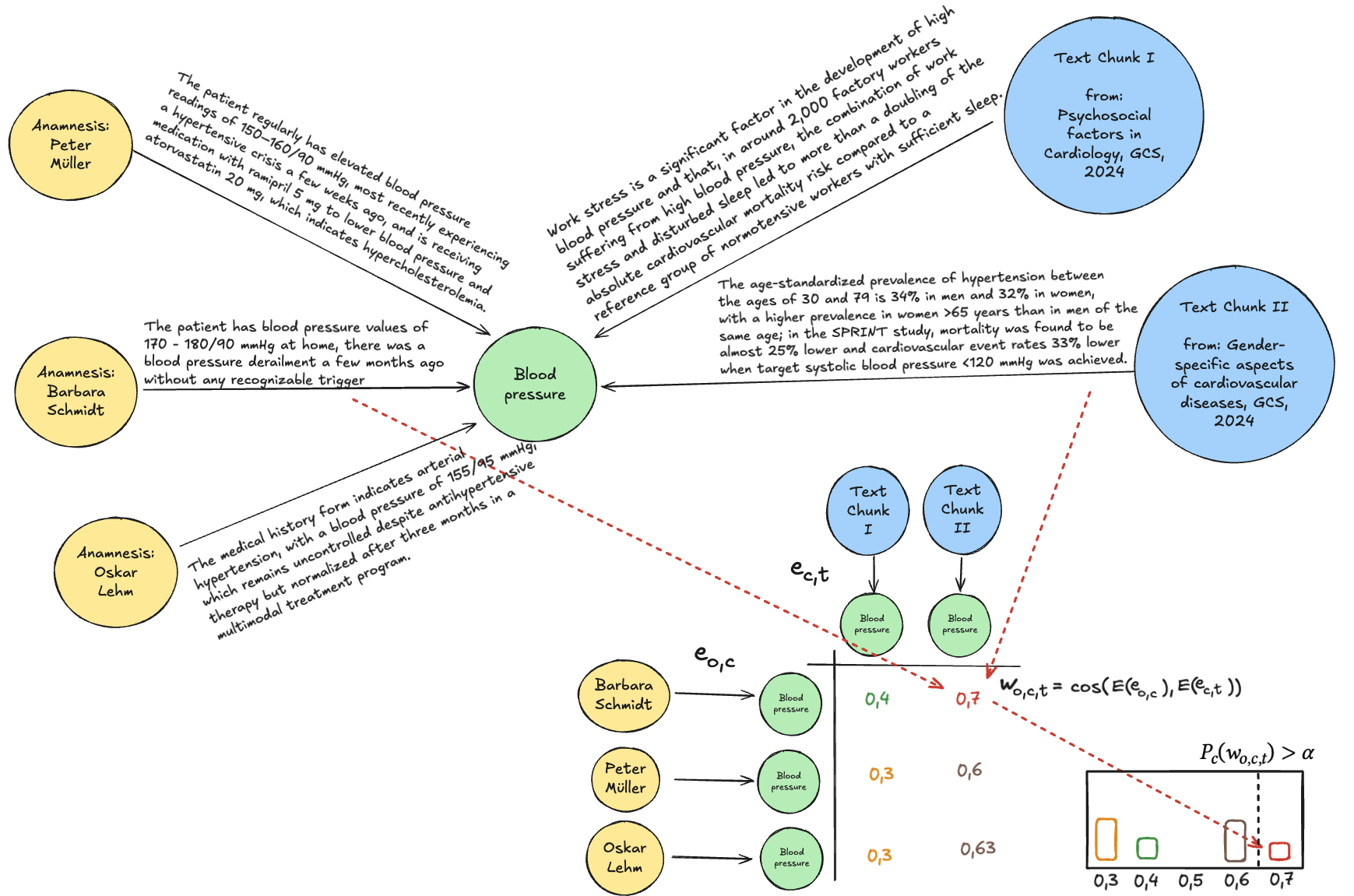}
\caption{Illustration of the derivation of $P_{c}$ via collection of all combinations $w_{o,c,t} = cos(E_{e_{o,c}}, E_{e_{c,t}})$ for all $o \in $ (Peter Mueller, Barbara Schmidt, Oskar Lehm) and $t \in $ (text chunk I, text chunk II), as illustrated in the respective table and exemplary histogram. E.g. assuming $w_{o={\text{Barbara Schmidt}},c,t={\text{text chunk II}}} = cos(E_{e_{o={\text{Barbara Schmidt}},c}}, E_{e_{c,t={\text{text chunk II}}}}) = 0.7$ with $c = \text{'blood pressure'}$, therefore, $P_c(w_{o={\text{Barbara Schmidt}},c,t={\text{text chunk II}}}) > \alpha$. Given the provided query to be related to object $o={\text{'Barbara Schmidt'}}$ would result in the selection of $t={\text{'text chunk II'}}$.} \label{fig_example}
\end{figure}

See figure~\ref{fig_example} for an example. $P_{c}$ is derived based on collecting all six $w_{o,c,t} = cos(E_{e_{o,c}}, E_{e_{c,t}})$ for all $o \in $ (Peter Mueller, Barbara Schmidt, Oskar Lehm) and $t \in $ (text chunk I, text chunk II). E.g. let us assume $w_{o = \text{Barbara Schmidt},c,t={\text{text chunk II}}} = cos(E_{e_{o = \text{Barbara Schmidt},c}}, E_{e_{c,t={\text{text chunk II}}}}) = 0.7$ with $c = \text{blood pressure}$, hence forming the top element of the empirical distribution (illustrated as histogram in figure~\ref{fig_example}). For the given example, $P_c(w_{o={\text{Barbara Schmidt}},c,t={\text{text chunk II}}}) > \alpha$, therefore, if the provided query would be related to the cross-examination of object $o={\text{'Barbara Schmidt'}}$, this would result in the selection of $t={\text{'text chunk II'}}$ as context for the prompt to answer the query. To this end, this fast, i.e. embedding based, pre-evaluation of potentially relevant text chunks based on applying some hyper-parameter $\alpha$ on $P_c$ not only serves as a pre-filter in order to avoid cluttering the language model context window but also as a measure to facilite scalability as the number of objects $o$ and text elements $t$ related to a concept $c$ increases.

\newpage

Further, in $G'_{o}$ two nodes $x_i, x_j$ are connected, if their respective $e_{c_i,t}$, $e_{c_j,t}$ form a first-order connection via a common $t$ in $G$, indicating that text chunk $t$ discusses both concepts $c_i$ and $c_j$ (see illustration in figure~\ref{fig_graph_transformation}). Hence, let's assume there exists both a connection $e_{o,c_i}$ and $e_{o, c_j}$ in $G$, while $P_{c_i}(w_{o,c_i,t}) > \alpha$ and $P_{c_j}(w_{o,c_j,t}) < \alpha$, i.e. $x_i = 1$ and $x_j = 0$. Then we can introduce a second, more lenient, condition-specific, interaction-based hyper-parameter $\beta$, e.g. $\beta = 0.5$, and evaluate $x_j = 1$ if $x_i = 1$ and $P_{c_j}(w_{o,c_j,t}) > \beta$ to indicate that the information about $c_j$ in $x_j$ might still be relevant w.r.t. $o$ - even at a lower absolute $w_{o,c_j,t}$ -  given its co-discussion in $t$ together with $c_i$. Note that this step is inspired by principles from Markov Random Fields ~\cite{koller2009, banf2017}, which implement a local independence assumption referred to as Markov Blanket ~\cite{koller2009}. 

A final prompt regarding the analysis of $o$ is constructed by: i) initially stating a basic query related to the cross-examination of a given complex object $o$; ii) followed by a sequential listing of the concept-specific summaries in $e_{o,c}$; iii) each separately followed the concept-specific summaries stored in all selected $e_{c,t}$, i.e. $e_{c,t}$ with $x = 1$, for all $(c, t)$ associated with $o$, along with document references. Note that the order of the information stored in all selected $e_{c,t}$ per $e_{o,c}$ is given by the existing lexical relationship structure in the graph between subsequent text chunks within and across sections of a document. 

\section{A first experiment}

\begin{figure}[h!]
\includegraphics[width=\textwidth]{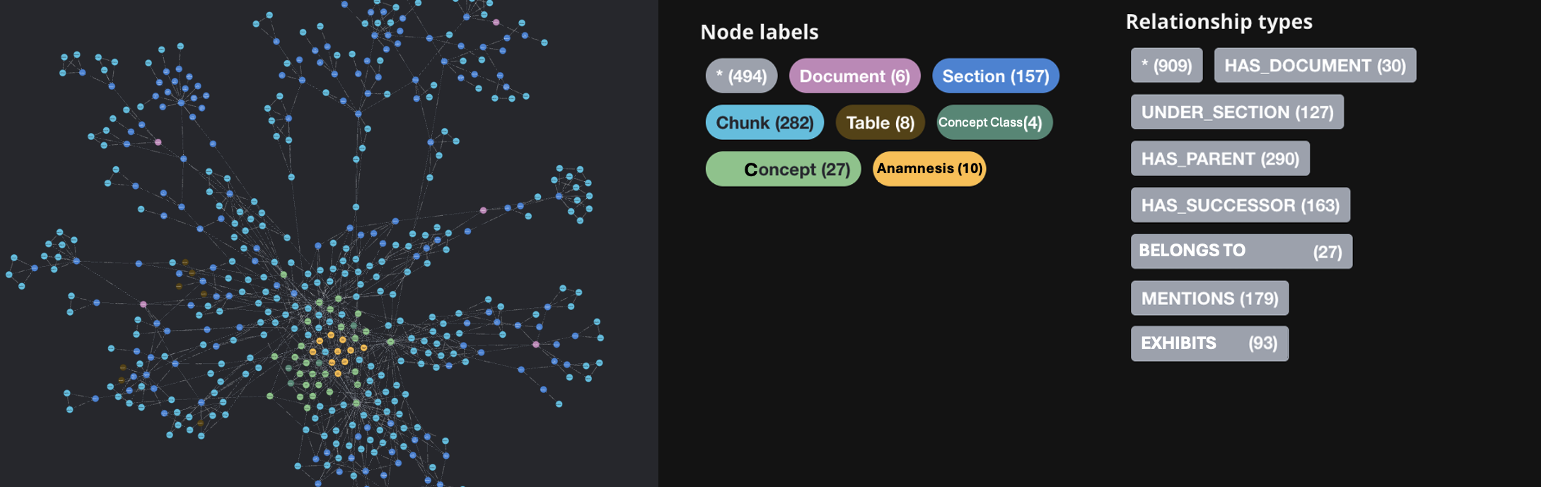}
\caption{The generated tripartite medical knowledge graph representation in Neo4j.} \label{fig_noe4j_medgraph}
\end{figure}

\begin{figure}[h!]
\includegraphics[width=\textwidth]{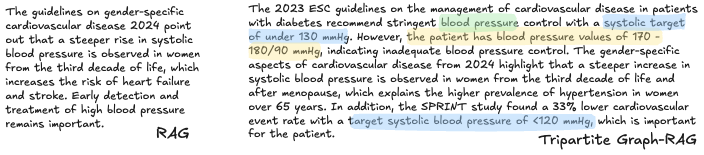}
\caption{In contrast to a generic response by RAG (left), Tripartite-GraphRAG (right) links finding (yellow) from a patient anamnesis to corresponding concept (green) and identifies two relevant, concept-specific source statements (blue) for examination.} \label{fig_comparison}
\end{figure}

Based on a healthcare use case, we construct a first experiment for the multi-faceted analyses of patient anamneses given a set of medical concepts given clinical guideline literature. We collect six guidelines from the European Cardiological Society on topics such as diabetes, pulmonary hypertension or gender-specific aspects with on average 31 ($\sigma=15,5$) sections and 47 ($\sigma=10$) text chunks $t$, each varying in length. In total 4 concept classes with 27 concepts $c$ related to diseases, symptoms, medication and risk factors were manually curated. As objects of investigation $o$ we task GPT4o to generate anamneses of fictitious patients (see example in figure~\ref{fig_concept_specific_summaries}). We use GPT4o-mini and Neo4j for the construction of the $o,c,t$ tripartite knowledge graph (see figure~\ref{fig_noe4j_medgraph}), while GPT4o is used for addressing the final query. Comparison of our approach is against a naive RAG setup \cite{gao2023retrieval}, devising similar queries, i.e. to take an anamnesis object $o$ as input and cross-examine all mentioned medical conditions given the provided literature. We investigate 3 anamnesis objects $o$, containing on average 11 ($\sigma=4$) concepts $c$. For Tripartite-GraphRAG, we set $\alpha = 0.9, \beta = 0.5$, resulting in prompt lengths of 3500 ($\sigma=1600$) tokens on average. For RAG, we ran three versions per each $o$ with varying similarity score minima resulting in prompt lengths ranging from approx. 3000 to 30000 tokens. Here, we are interested in the amount of recovered information, w.r.t to prompt length, which relates to the notion of information density~\cite{Aceves2024}. In comparison, we observe a significantly higher information density for Tripartite-GraphRAG, recovering on average 8 concepts at approx. 1500 - 5500 tokens, compared to RAG's on average 4 - 6 concepts at approx. 3000 - 31000 tokens. In addition, we observe that prompt configuration and arrangement via Tripartite-GraphRAG not only influences the number of concepts recovered, but also how thoroughly concepts are cross-examined. Taking the following finding in a patient's anamnesis as an example, i.e. \textit{"the patient measures at home blood pressure values of 170 - 180/90 mmHg"}, RAG - in all of our experiments - selected only one, if any, of the two relevant source statements, let alone being unable to compare them properly. As a result RAG, while not necessarily being wrong, often produced generic results (see figure~\ref{fig_comparison}, left). In contrast, Tripartite-GraphRAG directly linked the finding to the corresponding concept $c = \text{blood pressure}$ and identified appropriate source statements for examination (see figure~\ref{fig_comparison}, right).

\newpage

\section{Discussion: current limitations and ongoing research}

Our Tripartite-GraphRAG approach demonstrates a novel paradigm for combining LLMs with knowledge graphs through concept-anchored pre-analysis and graph-based prompt optimization. The tripartite structure offers several advantages: it implements information-preserving pre-compression, enables concept-specific relevance estimation grounded in statistics, and avoids common challenges such as entity resolution and deduplication when extending the knowledge base. By formulating prompt creation as an unsupervised node classification problem, we leverage the graph structure to optimize information arrangement and coverage systematically.

Despite these promising results, our current work has several limitations that warrant discussion and guide our ongoing research. First, the manual curation of the domain-specific ontology represents a significant bottleneck in our current implementation. While the conceptual design of the tripartite graph as a supporting structure is invariant to the method of ontology creation, the manual process limits scalability and may introduce biases in concept selection and hierarchical organization. Second, our experimental evaluation remains primarily anecdotal, focusing on a limited healthcare use case with only three patient anamneses and manually generated fictitious data. The comparison baseline is limited to naive RAG implementations, without evaluation against more sophisticated retrieval-augmented approaches ~\cite{sarthi2024raptor, rackauckas2024ragfusion, gao2024modular}. Third, the hyperparameters $\alpha$ and $\beta$ were selected empirically without systematic optimization, potentially leaving room for performance improvements due to domain-specific inference and learning approaches.

Several promising avenues emerge for future research. Most critically, we plan to integrate established domain-specific ontologies to enable advanced reasoning capabilities over the knowledge graph. For the medical domain, this includes incorporating standardized ontologies such as SNOMED CT~\cite{donnelly2006snomed}, the Unified Medical Language System (UMLS)~\cite{bodenreider2004unified}, or the Human Phenotype Ontology (HPO)~\cite{robinson2008human}. This integration will require developing more fine-grained automated inference mechanisms to determine which ontology concept nodes should be utilized for domain summarization distribution generation. To address the limitations in our experimental evaluation, we plan to conduct comprehensive benchmarking against state-of-the-art methods and datasets. We will expand our experiments to include standardized medical question-answering datasets such as MedQA~\cite{jin2021disease} and PubMedQA~\cite{jin2019pubmedqa}, enabling more rigorous quantitative comparisons. Additionally, we aim to evaluate our approach w.r.t. recent work integrating ontological reasoning with LLMs to enhance knowledge-grounded generation \cite{pan2024unifying, hohenecker2020ontology, chen2023owl2vec, ye2024llm} which has demonstrated that incorporating ontologies into neural models significantly improves reasoning over incomplete knowledge graphs, in particular for medical applications \cite{kommineni2024human, liu2024ontoglm}. 

Finally, we plan to explore the integration of our approach with emerging techniques in trustworthy AI, particularly focusing on improving provenance tracking and explainability of the reasoning process. This includes developing visualization tools to make the concept-based retrieval and reasoning process more transparent to end-users, particularly important in critical domains such as healthcare and industrial automation.

%%
%% Define the bibliography file to be used
% \bibliography{sample-ceur}

%

\end{document}